\documentclass[11pt]{article}
\usepackage{eacl2017}
\usepackage{times}
\usepackage{graphicx}
\usepackage{url}
\usepackage{latexsym}
\usepackage{xcolor}
\usepackage{tipa}
\usepackage{amsmath}
\usepackage{comment}

\usepackage{floatrow}  

\pdfoutput=1

\eaclfinalcopy 

\title{Phonological (un)certainty weights lexical activation}

\author{
    Laura Gwilliams\textsuperscript{1} \qquad David Poeppel\textsuperscript{1}\qquad Alec Marantz\textsuperscript{1,2} \qquad Tal Linzen\textsuperscript{3}\\
    \textsuperscript{1}Department of Psychology, New York University\\
    \textsuperscript{2}Department of Linguistics, New York University\\
    \qquad \textsuperscript{3}Department of Cognitive Science, Johns Hopkins University\\
    {\tt \{leg5, dp101, marantz\}@nyu.edu \qquad tal.linzen@jhu.edu}
}

\date{}

\begin{document}
\maketitle
\begin{abstract}
Spoken word recognition involves at least two basic computations. First is matching acoustic input to phonological categories (e.g. \textipa{/b/, /p/, /d/}…). Second is activating words consistent with those phonological categories. Here we test the hypothesis that the listener's probability distribution over lexical items is weighted by the outcome of both computations: uncertainty about phonological discretisation and the frequency of the selected word(s). To test this, we record neural responses in auditory cortex using magneto-encephalography, and model this activity as a function of the size and relative activation of lexical candidates. Our findings indicate that towards the beginning of a word, the processing system indeed weights lexical candidates by both phonological certainty and lexical frequency; however, later into the word, activation is weighted by frequency alone.
\end{abstract}


\section{Introduction}
There is mounting evidence for the predictive nature of language comprehension. Response times and neural activity are reduced in response to more predictable linguistic input. This indicates that the brain forms probabilistic hypotheses about current and future linguistic content, which manifest in expectations of phonemes, morphemes, words and syntactic structures \cite{connolly1994event,lau2006role,lau2008cortical,ettinger2014role,gwilliams2015non}.

In speech comprehension, the brain's task is to correctly determine a word's identity as quickly as possible. It is not optimal to always wait until word ending, because the target may be correctly identifiable earlier. For example, after hearing \textit{hippopotamu-} the final \textipa{/s/} provides very little additional information. Indeed, one could even stop at \textit{hippot-} and still identify the target word correctly most of the time.\footnote{ Note that \textit{hippopotomonstrosesquippedaliophobia} (`fear of long words') and \textit{hippopotas} (`a ground-type Pokemon') are also possible lexical items but much less frequent than the target in this case, so less likely to be selected.} 

How is this done? Upon hearing the beginning of a lexical item, the brain activates the cohort of words that are consistent with the acoustic signal. Words in the cohort are activated relative to their match to the phoneme sequence and frequency of occurrence. With each subsequent phoneme, the cohort is reduced as items cease to be consistent with the provided input, until one item prevails (see Figure 1). This process is consistent with the highly influential cohort model of spoken word recognition  \cite{marslen1978processing,marslen1987functional}, and has been associated with activity in left superior temporal gyrus (STG) \cite{gagnepain2012temporal,ettinger2014role,gwilliams2015non}.

In practice though, phoneme identity is often uncertain: the acoustic signal may be consistent with both a [\textipa{b}] and a [\textipa{p}], for example. This phonetic uncertainty, and its effect on lexical activation, is not addressed by the cohort model. However, there is evidence suggesting that phonetic uncertainty affects lexical and sentential processing \cite{connine1991effects,mcmurray2009within,bicknell2015listeners}.




\begin{figure*}[h]
\includegraphics[width=\linewidth]{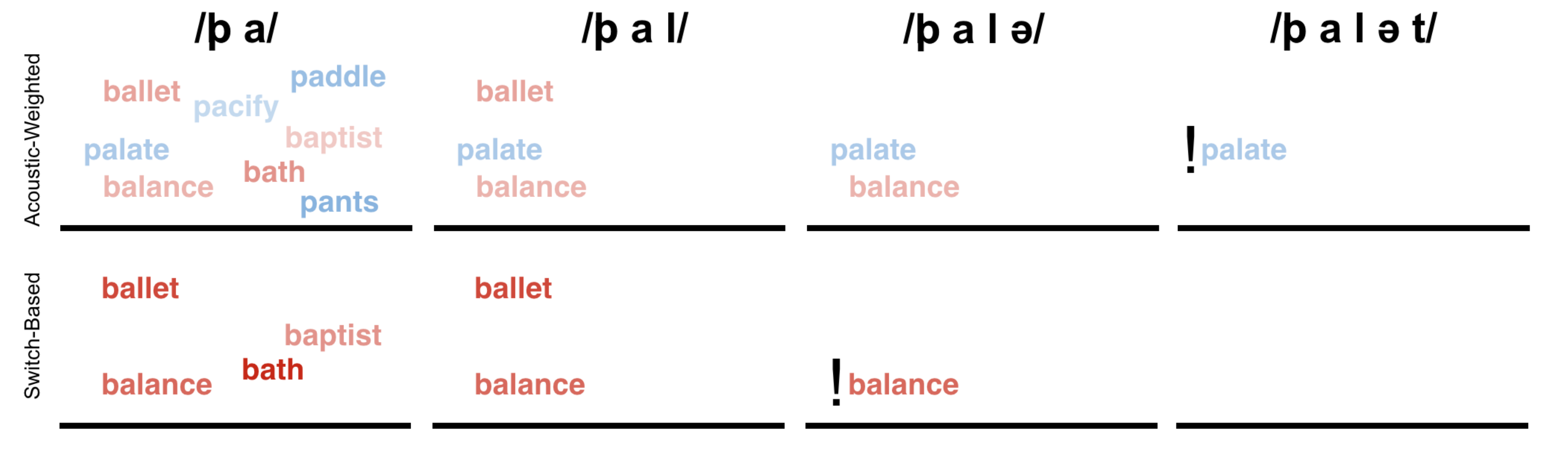}
\caption{Schematic depiction of cohort activation under each of the two models, for the first five phonemes of the word \textit{palate}. The onset b-p symbol represents that the onset phoneme was 75\% consistent with a /\textipa{b}/ and 25\% consistent with a /\textipa{p}/. Transparency reflects relative word activation. Note that the change in transparency between the two accounts reflects the actual probabilities predicted by each model --- because there are more words activated in the Acoustic-Weighted account, less normalised probability is assigned to each item.}
\end{figure*}


Here we build upon this previous work in order to understand the neural computations underlying lexical activation, in service to spoken word recognition. Concretely, we ask: How does fine-grained acoustic information (below the phonological level) serve to activate lexical hypotheses and estimate their probabilities? Can this integration between phonological and lexical levels of description be read out from the STG? 

To address these questions, we model neural responses in STG, time-locked to each phoneme in a word, as a function of two computational models. One model assumes that the activation of a lexical candidate is gradiently weighted by the acoustic evidence in favour of that candidate: e.g., \textit{balloon} is activated in proportion to how /\textipa{b}/-like the initial sound of the word was, even if that sound was more likely to represent a different phoneme (e.g., /\textipa{p}/). We refer to this model, in which phonetic uncertainty is carried over to the word recognition process, as the \textbf{acoustic-weighted} model. The other model assumes that acoustic information serves as a  switch: a lexical item is either fully activated or not activated at all, as a result of a discrete decision made at the phonetic level. This model, which we refer to as the \textbf{switch-based} model, is most consistent with the traditional cohort model —-- the system commits to whichever phoneme is more likely, and this is used to form predictions at the lexical level (see Figure 1). A subset of the data reported here are also published in \newcite{gwilliams2017spoken}.


\section{Summary of human data}

\begin{figure}[!b]
\includegraphics[width=\linewidth]{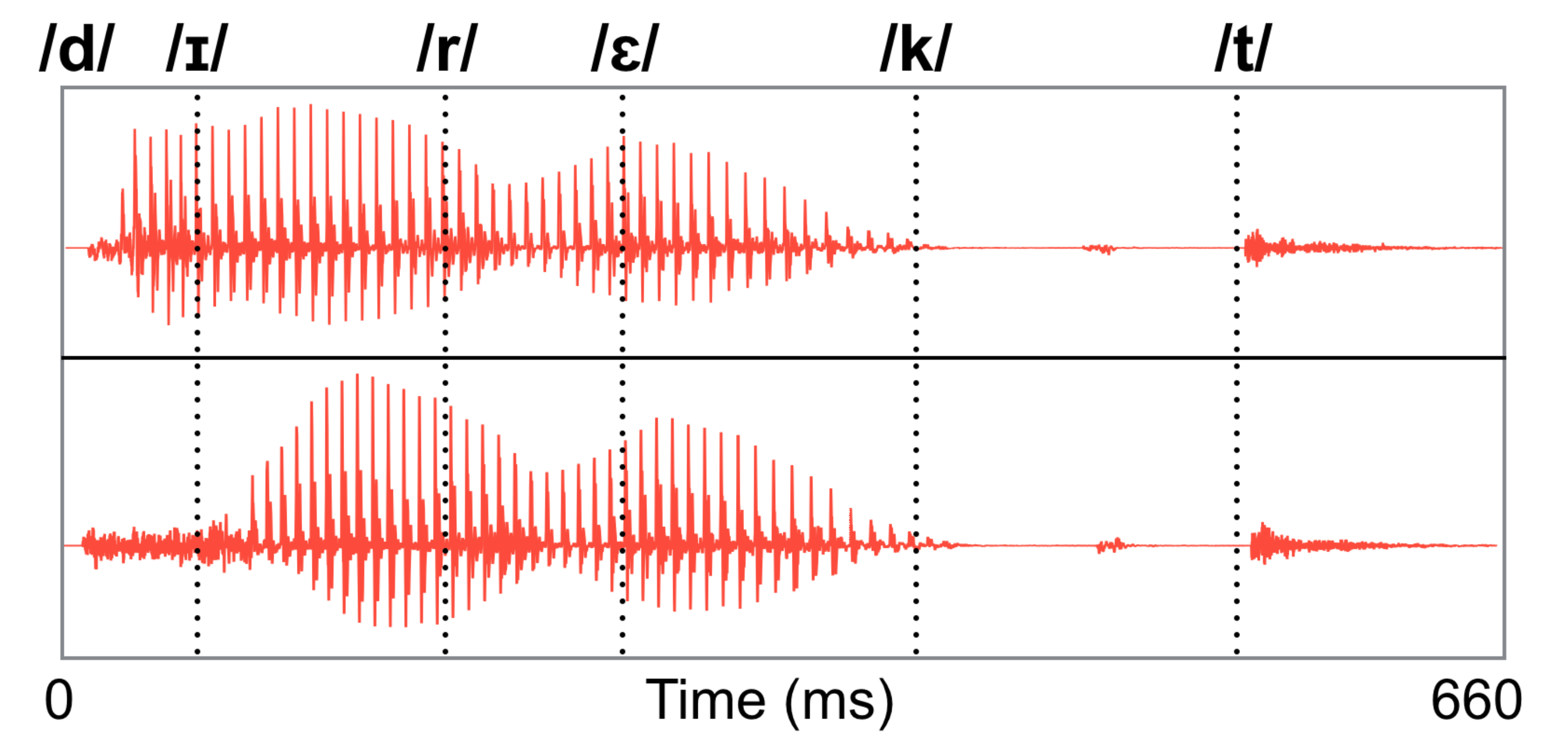}
\caption{\label{waveform}Waveforms of example endpoints of a lexical continuum. The word \textit{direct} is above, and the non-word \textit{tirect} is below. Dashed lines correspond to the timing of each phoneme onset.}
\end{figure}


\subsection{Materials}
Word pairs were selected such that, apart from the first phoneme, there was an identical phoneme sequence until a point of disambiguation. For example, \textit{palate} and \textit{balance} share their second, third and fourth  phonemes ([\textipa{\ae}], \textipa{[l]} and [\textipa{@}], respectively), and diverge on the fifth (\textipa{[t]} vs. \textipa{[n]}). We selected 103 word pairs with this property. The onset of each word was either a voiced (\textipa{d, b, g}) or voiceless (\textipa{t, p, k}) plosive. A native English speaker was recorded saying each of these 206 words in isolation. The onset of each word was morphed along one phonetic feature, using the TANDEM-STRAIGHT software to create a 11-step continuum between word (e.g., \textit{direct}) and non-word (e.g., \textit{tirect}) (see Figure \ref{waveform}). The 11-step acoustic continuum was then re-sampled to form a 5-step perceptually defined continuum, based on the proportion of selections in a behavioural pre-test.


\begin{figure*}[!t]
\includegraphics[width=\linewidth]{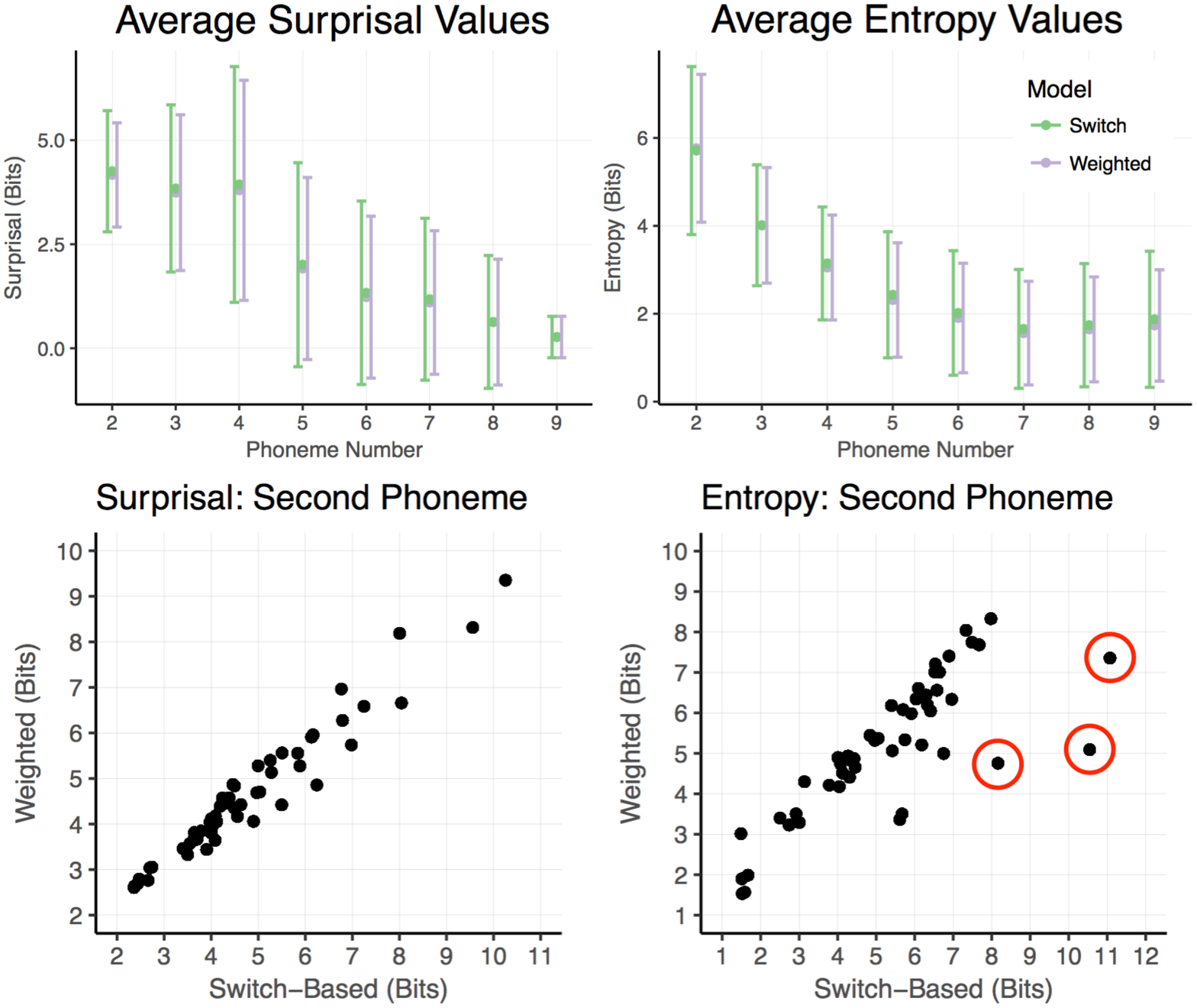}
\caption{Top: Average surprisal and entropy values at each phoneme along the word. Note that not all words are 9 phonemes long, so phonemes at longer latencies contain fewer entries. Error bars represent one standard deviation from the mean. Bottom: Correlation between the two models' surprisal values and the two models' entropy values, at the second phoneme. Red circles highlight the outliers  \textit{topography}, \textit{tirade} and \textit{casino} (from right to left).}
\end{figure*}

\subsection{MEG experiment}
Native English participants ($n = 25$) listened to each of the 103 $\times$ 5 words in isolation, and in 20\% of trials (randomly distributed) made an auditory-to-visual word matching judgment. 

While completing the task, neural responses were recorded using a 208-sensor KIT magnetoencephalography (MEG) system. Data were sampled at 1000 Hz, which provided a measure of neural activity at each millisecond. In order to test responses to specific phonemes in a word, the data were cut into a series of 700 ms epochs, where the time at 0 ms corresponds to the onset of a phoneme. Note that the phonemes were shorter than 700 ms, so the epochs overlapped in time. The activity recorded from MEG sensors was localised using MNE-Python software \cite{gramfort2014mne}, and averaged over the left STG. This provided one datapoint per millisecond (700) per phoneme (4370) per participant (25).

\section{Modeling of MEG data}
The variables of interest were entropy and surprisal. Entropy quantifies uncertainty about the resulting lexical item. For switch-based entropy we followed the typical calculation, which assumes that only the words whose phonemes are \textit{most} consistent with the acoustics are included in the activated cohort (e.g., only the \textit{b}-onset words):

\begin{equation}\label{entropy}
  -\sum\limits_{w \in C}P(w|C) log_2P(w|C)
\end{equation}

\noindent where $C$ is the set of all words consistent with the heard prefix, and 

\begin{equation}
	P(w|C) = \frac{f(w)}{\sum\limits_{w \in C}f(w)}
\end{equation} 

\noindent where $f(w)$ is the frequency of the word $w$.

For acoustic-weighted entropy, the cohort is made up of two sub-cohorts, $C_a$ and $C_b$, one for each of the possible word-initial phonemes (e.g.,  /b/ and /p/). The conditional probabilities of the words in each sub-cohort $C_a$ and $C_b$ were weighted by the probabilities of each possible onset phoneme given the acoustic signal $A$, which we derived from the behavioural pretest:

\begin{equation}
	\begin{array}{lll}
  		P(w|C, A) & = & P(w|C_a)P(\varphi_a|A) +\\
        & & P(w|C_b)P(\varphi_b|A)
  	\end{array}
\end{equation}

\noindent where $\varphi_a$ and $\varphi_b$ are the two phonemes consistent with the acoustic signal $A$. These acoustic-weighted measures of word frequency and cohort frequency were then used in the typical entropy calculation given in Equation \ref{entropy}. We note that switch-based entropy can be understood as the result of rounding the acoustic weighting terms $P(\varphi_a|A)$ and $P(\varphi_b|A)$ to their nearest integer (either 1 or 0; see Figure 1).



Surprisal quantifies how expected the current phoneme $\varphi_t$ is given the prior phonemes $\varphi_1,\ldots,\varphi_{t-1}$:

\begin{equation}
	\begin{aligned}
  -log_2\frac{f(\varphi_1,\ldots,\varphi_t)}{f(\varphi_1,\ldots,\varphi_{t-1})}
  	\end{aligned}
\end{equation}

\noindent where $f(\varphi_1,\ldots,\varphi_t)$ denotes the summed frequency of all words that start with the phoneme sequence $\varphi_1,\ldots,\varphi_t$.


For switch-based surprisal, the conditional probability is calculated from the cohort of words most consistent with the acoustics at onset: e.g. the \textit{b}-onset words. To calculate acoustic-weighted surprisal, we estimate the conditional probability separately for each cohort of words ($_a$, $_b$), and then scale each conditional probability by an acoustic weighting term and a lexical weighting term:

\begin{equation}
	\begin{aligned}
  -log_2\left(P(\varphi_a|A)\frac{f(\varphi_a,\varphi_2,\ldots,\varphi_t)}  {f(\varphi_a,\varphi_2,\ldots,\varphi_{t-1})} Q_a^t +\right.\\
  \left.P(\varphi_b|A)\frac{f(\varphi_b,\varphi_2,\ldots,\varphi_t)}  {f(\varphi_b,\varphi_2,\ldots,\varphi_{t-1})} Q_b^t\right)
  	\end{aligned}
\end{equation}


\noindent where

\begin{equation}
	Q_a^t = \frac{f(\varphi_a,\varphi_2,\ldots,\varphi_t)}{f(\varphi_a,\varphi_2,\ldots,\varphi_t) + f(\varphi_b,\varphi_2,\ldots,\varphi_t)}
\end{equation}

The $Q$ lexical weighting is the probability of the observed sequence, given a cohort that contains both $\varphi_a$ and $\varphi_b$-onset words. The acoustic weighting is the same as described above.

In all, this surprisal value is calculated by estimating the probability of each phoneme $\varphi_a$, $\varphi_b$ given i) acoustics; ii) preceding phonemes; iii) probability of the sequence given a joint cohort. The probability of each phoneme is then summed before taking the negative logarithm. This derives an overall surprisal of the sound, given the phonological categories it could realise.

For all of these calculations, word frequencies were extracted from the English Lexicon Project \cite{balota2007english}.

As shown in Figure 3, the surprisal and entropy calculations from the two models were highly correlated. This is because here we are re-analysing a dataset that was designed and collected for other reasons. In future work we plan to design materials that maximally distinguish switch-based and acoustic-based accounts. Our results stand in as a first approximation that can (and should) be built upon.


\section{Results}

\begin{figure*}[h]
\centering
\includegraphics[width=0.8\linewidth]{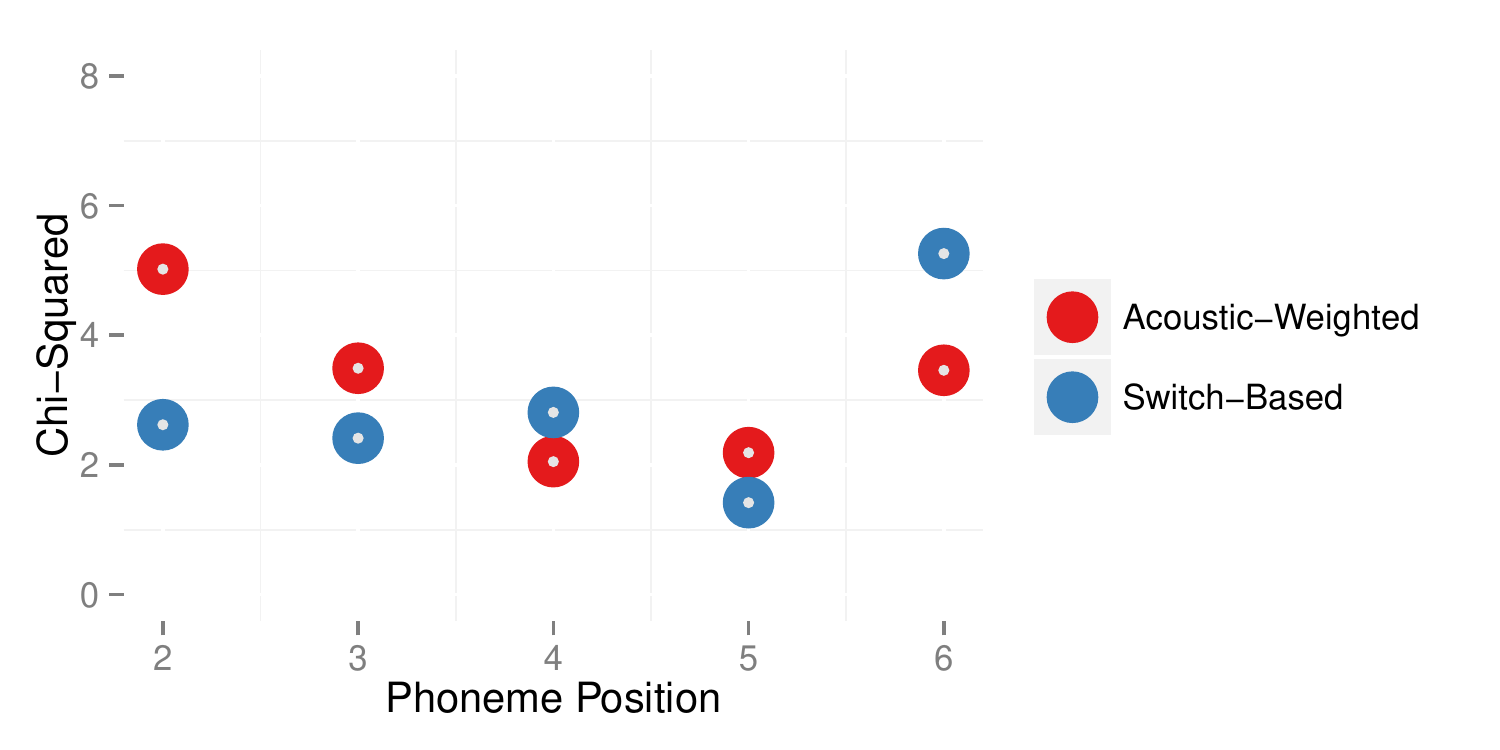}
\caption{Reduction in linear mixed-effects model log-likelihood resulting from excluding acoustic-weighted surprisal and entropy (in red) or switch-based surprisal and entropy (in blue); higher values indicate that the predictors increase model fit more. The dependent measure was activity averaged from 200-250 ms in STG, time-locked to phonemes along the length of the words.}
\end{figure*}

The dependent measure was activation of left STG, averaged between 200-250 ms after phoneme onset, a time window determined based on \newcite{ettinger2014role}. This activity was modelled time-locked to each phoneme along the length the word, but we primarily focused on the second (mean post-onset latency = 87 ms; SD = 25 ms, 4021 observations) and the sixth phonemes (mean post-onset latency = 411 ms; SD = 78 ms, 3264 observations). This was because they included a similar number of trials in each model comparison, while also ensuring substantial differences in latency from word onset. Reported results were corrected for multiple comparisons over all six phoneme positions using Bonferroni correction. Only responses to partially ambiguous trials were included (0.25 and 0.75), because this is where the predictions of acoustic-weighted and switch-based models are most distinct.

We evaluated the fit of the predictions of each model to the neural measurement using a linear mixed effects  model. The full model contained switch-based and acoustic-weighted surprisal, switch-based and acoustic-weighted entropy, phoneme latency, trial number, block number, stimulus amplitude of the first 30 ms, phoneme pair and ambiguity as fixed effects. By-subject slopes were included for all entropy and surprisal predictors. This full model was compared to a model where either acoustic-entropy and surprisal, or switch-based entropy and surprisal, were removed as fixed effects (but remained as by-subject slopes). This gave a statistical assessment of the amount of variance the acoustic-weighted and switch-based models were accounting for.

At the second phoneme, the acoustic-weighted variables explained a significant amount of variance ($\chi^2 = 5.02$, $p = .025$), whereas the switch-based variables did not ($\chi^2 = 2.62$, $p = .1$). At the third phoneme, the acoustic-weighted variables were marginally significant ($\chi^2 = 3.49$, $p = .061$), the switch-based variables were not ($\chi^2 = 2.41$, $p = .12$). At the fourth phoneme, neither model was significant: Acoustic weighted ($\chi^2 = 2.05$, $p = .15$) or switch-based ($\chi^2 = 2.81$, $p = .094$). The same was true at the fifth phoneme: Acoustic weighted ($\chi^2 = 2.19$, $p = .14$), switch-based ($\chi^2 = 1.42$, $p = .23$). At the sixth phoneme, we observe the opposite effect from the second phoneme position: the switch-based variables explained a significant amount of variance ($\chi^2 = 5.26$, $p = .022$) and the acoustic-weighted variables had only marginal explanatory power ($\chi^2 = 3.46$, $p = .06$). These results are displayed in Figure 4.

\section{Discussion}

We have found evidence that the brain uses fine-grained acoustic information to weight lexical predictions in spoken word recognition. At the beginning of a word, lexical hypotheses are activated in proportion to the bottom-up acoustic evidence; towards the end, acoustic evidence acts as a switch-like function, to either fully activate or deactivate the word, bounded by its frequency of occurrence. This finding has two primary implications.

First, it suggests that the system does not wait until phonological categories have been disambiguated before activating lexical items. Rather, uncertainty about phonological classification is used to modulate higher level processes, ensuring that phonological discretisation is not a bottleneck in activating lexical items. This supports interactive models of speech processing, because it suggests that the output of one stage does not need to be determined before initiating the following. In particular, this finding is inconsistent with the Cohort model of speech perception \cite{marslen1978processing}, which assumes that the system first commits to the most likely phoneme before making lexical predictions.

Second, it suggests that the same processing strategy is not heuristically applied in all situations. Rather, phonological information appears to be used more when processing the beginning of a word than the end. There are two explanations for this. This could reflect that the system commits to a particular phonological category after a given delay period, and so the phonological weights used by the system converge to a stable decision point. Or perhaps lexical frequency becomes more informative as the size of the cohort decreases, and so phonological detail is given less predictive power by the processing system. A simple way to tease these alternatives apart in future work is to manipulate the ambiguity of phonemes within a word, not just in initial position. The former would predict that acoustic evidence is used in close proximity to the ambiguous sound, regardless of its position in the word; the latter would predict that acoustic evidence is used more at the beginning of the word, regardless of the position of the ambiguous sound.






\bibliography{phoneme_surprisal_cmcl}
\bibliographystyle{eacl2017}

\end{document}